\documentclass[11pt]{article}
\usepackage[margin=1in]{geometry}
\usepackage[T1]{fontenc}
\usepackage{lmodern}
\usepackage{microtype}
\usepackage{amsmath, amssymb}
\usepackage{graphicx}
\usepackage{booktabs}
\usepackage{siunitx}
\usepackage[authoryear]{natbib}
\usepackage[hidelinks]{hyperref}
\usepackage[nameinlink,noabbrev]{cleveref}
\usepackage{authblk}
\usepackage{fancyhdr}

\title{Why Any-Order Autoregressive Models Need Two-Stream Attention: A Structural–Semantic Tradeoff}
\author[ ]{Patrick Pynadath,}
\author[ ]{Ruqi Zhang}
\affil[ ]{Department of Computer Science, Purdue University}
\date{\today}

\fancypagestyle{firstpage}{
  \fancyhf{}
  \fancyfoot[L]{\textit{Correspondence: \href{mailto:ppynadat@purdue.edu}{ppynadat@purdue.edu}}}

}

\begin{document}
\maketitle
\thispagestyle{firstpage}

\begin{abstract}
\noindent
Any-order autoregressive models (AO-ARMs) offer a promising path toward efficient masked diffusion by enabling native key-value caching, but competitive performance has so far required two-stream attention, typically motivated as a means of decoupling token content from position. In this work, we argue that two-stream attention may be serving a more subtle role. We identify a \emph{structural-semantic} tradeoff in any-order generation: the hidden representation at each step must simultaneously attend to semantically informative tokens for prediction and structurally recent tokens for summarization, objectives that compete for attention capacity in a single stream but can specialize across two streams. To isolate this tradeoff from position-content separation, we propose \emph{Decoupled RoPE}, a modification to rotary position embeddings that provides target position information without revealing target content. Decoupled RoPE performs competitively at short sequence lengths—where semantic and structural proximity coincide—but degrades as sequence length increases and the two orderings diverge. These results suggest that the success of two-stream attention stems not merely from separating position from content, but from circumventing the deeper structural-semantic tradeoff inherent to any-order generation.
\end{abstract}

\section{Introduction}
Masked diffusion models (MDMs) have emerged as a compelling alternative to autoregressive generation, offering the ability to generate multiple tokens in parallel by framing sequence generation as a diffusion process over discrete masked states \citep{sahoo2024simple, shi2024simplified}. 
However, a key practical limitation of MDMs is the lack of native key-value (KV) caching: because they are typically parameterized as bidirectional encoders, each denoising step requires recomputing attention over the entire sequence, incurring substantial computational cost at inference time.

A natural remedy follows from the well-known equivalence between masked diffusion and any-order autoregression models (AO-ARMs) \citep{ou2024your, sahoo2025esoteric}: rather than parameterizing the model as a bidirectional encoder, one can use a standard causal decoder, which directly supports KV caching. 
This approach has attracted significant recent interest, as it could allow MDMs to fully realize the efficiency gains of parallel generation without sacrificing inference speed \citep{du2026autoregressive, guo2025reviving, karami2026auto, sahoo2025esoteric, xue2025any}.

Interestingly, many of these works leverage two stream attention, introduced in \citet{yang2019xlnet} as a means of allowing the model to see the position it should predict without leaking the content of that position. Two stream attention is typically motivated as a way to provide the model with target position information without revealing the content at that position. 

In this work, we revisit this assumption and ask a more fundamental question: \emph{is separating position from content sufficient for competitive any-order autoregression?} 
We argue that it is not, and that single-stream causal attention faces a deeper obstacle — a \textbf{structural-semantic tradeoff} — that position-content separation alone cannot resolve. 
At each generation step, the hidden representation must serve two purposes: (1) summarizing the most relevant context for predicting the next token, which favors attending to \emph{semantically} nearby positions, and (2) aggregating the most complete information about the sequence so far, which favors attending to \emph{structurally} recent positions — those that have seen the most context. 
In left-to-right autoregression, these two objectives are naturally aligned. In any-order generation, they generally conflict, forcing a single attention mechanism to compromise between them.

To make this case, we propose \textbf{Decoupled RoPE}, a modification to rotary position embeddings that cleanly encodes target position information without leaking target content. Decoupled RoPE serves as a controlled probe: if position-content separation were the primary bottleneck, this approach should perform competitively with masked diffusion. We show that it does not. 
Decoupled RoPE is competitive at short sequence lengths — where semantic and structural proximity tend to coincide — but degrades as sequence length increases and the two orderings diverge. 
This pattern of failure is consistent with the structural-semantic tradeoff, and suggests that the empirical success of two stream attention in prior work may stem not merely from separating position from content, but from circumventing this deeper tension.

\section{Background}

\subsection{Any-Order Autoregression}

Masked diffusion models generate sequences by iteratively unmasking tokens from a fully masked state, where each step transitions a subset of positions from masked to clean \citep{sahoo2024simple, shi2024simplified, ou2024your}. It is well known that this process is equivalent to any-order autoregression models (AO-ARMs): rather than generating tokens left-to-right, the model generates tokens according to a randomly sampled permutation $\pi$ of the sequence positions \citep{sahoo2025esoteric, ou2024your}. Under this equivalence, each denoising step corresponds to autoregressively predicting the next token in the permuted order, conditioned on all previously generated tokens.

This equivalence suggests a natural reparameterization: instead of using a bidirectional encoder, one can use a causal decoder that generates tokens according to the permuted order $\pi$. Given a sequence $\mathbf{x} = (x_1, \dots, x_n)$ and a uniformly sampled permutation $\pi$, the AO-ARM objective decomposes the joint likelihood as:
\begin{equation}
p(\mathbf{x}) = \prod_{t=1}^{n} p(x_{\pi(t)} \mid x_{\pi(1)}, \dots, x_{\pi(t-1)})
\end{equation}
A causal decoder can model this directly by processing the tokens in the order specified by $\pi$, with a standard causal attention mask ensuring that position $t$ in the structural order attends only to positions $1, \dots, t-1$. Critically, this enables native key-value caching, since each new token only needs to attend to previously generated tokens --- unlike bidirectional encoders, which must recompute attention over the entire sequence at each step.

However, a key challenge arises: at structural position $t$, the model must predict the token $x_{\pi(t+1)}$, which occupies some semantic position $\pi(t+1)$ in the original sequence. For this prediction to be well-informed, the model needs access to the \emph{semantic} position of the target --- that is, where in the original sentence the next token belongs --- without observing the content at that position. In standard left-to-right autoregression this is trivial, since the next structural position and the next semantic position coincide. In any-order generation, they generally do not.

\subsection{Two Stream Attention}

Prior works on any-order autoregression have addressed the position-content separation problem through two stream attention, introduced by \citet{yang2019xlnet}. Two stream attention maintains two parallel sets of hidden representations: a \emph{content stream}, which encodes the content of all tokens generated so far, and a \emph{query stream}, which has access to the target position but not the target content. 
At each layer, both streams perform attention over the previous positions, but the query stream excludes the content of the position it is predicting. This allows the model to condition its prediction on where the target belongs in the original sequence without leaking what token occupies that position.

Several recent works on any-order and any-subset autoregressive models have adopted two stream attention as a core architectural component \citep{du2026autoregressive, guo2025reviving, karami2026auto, xue2025any}, and it appears to be necessary for competitive performance. Two stream attention is typically motivated by the position-content separation argument above. 
In this note, we argue that it may be addressing a deeper issue, which we describe in the next section.
\section{The Structural-Semantic Tradeoff}

In a causal decoder, the hidden representation at each position must simultaneously accomplish two tasks: (1) \emph{predicting} the token at the next position, and (2) \emph{summarizing} the sequence generated so far so that downstream positions can build on it. Standard attention must distribute its capacity across previous positions to serve both goals, and the issue arises when these goals favor attending to different positions.

To formalize this tension, we distinguish between the two orderings introduced in the previous section. The \textbf{semantic order} is the natural order in which the tokens appear in the original sequence $(1, 2, \dots, n)$. The \textbf{structural order} is the generation order defined by the random permutation $\pi$. We posit two assumptions about how attention should ideally be distributed under each ordering:

\begin{enumerate}
    \item \textbf{Semantic locality}: for a given token, the most informative context tokens tend to be those that are nearby in semantic order. This reflects the well-documented locality bias of natural language --- nearby words are typically more relevant for prediction than distant ones.
    \item \textbf{Structural locality}: for a given step in generation, the most complete summary of the sequence so far is concentrated in the hidden representation of the immediately preceding structural position, since it has attended to all earlier positions in the generation order.
\end{enumerate}

In left-to-right autoregression, these two notions of locality are perfectly aligned: the structurally preceding position is also semantically adjacent, so attending to structurally recent positions simultaneously attends to semantically relevant ones. In any-order generation, this alignment breaks down. A random permutation $\pi$ scatters semantically adjacent tokens across arbitrary structural positions, so the positions that are most informative for \emph{prediction} (semantically nearby) are generally not the same as the positions that are most informative for \emph{summarization} (structurally recent).

\paragraph{Example.} Consider the sentence:
\begin{center}
    \texttt{The orange cat was very fluffy.}
\end{center}
Suppose the generation order is: \texttt{cat}, \texttt{fluffy}, \texttt{was}, \texttt{the}, \texttt{very}, and the model must now predict \texttt{orange}. The tokens \texttt{cat} and \texttt{fluffy} are semantically closest to \texttt{orange} and are likely the most informative for predicting it. However, the structurally most recent position --- \texttt{very} --- carries the most complete summary of the sequence, having attended to all four previously generated tokens. A single attention mechanism must trade off between these two objectives: increasing attention on semantically informative positions necessarily decreases attention on structurally complete ones, and vice versa.

\paragraph{Dependence on sequence length.} The severity of this tradeoff depends on sequence length. In short sequences, a randomly sampled permutation tends to place semantically nearby tokens close together in structural order as well, simply because there are fewer positions to scatter them across. As sequence length grows, the expected structural distance between semantically adjacent tokens increases, and the two orderings become increasingly misaligned. This suggests that single-stream approaches may perform reasonably at short sequence lengths but degrade as sequences grow longer --- a prediction we test empirically in \Cref{sec:experiments}.
\section{Decoupled RoPE}

To isolate the structural-semantic tradeoff from the simpler problem of position-content separation, we need a method that cleanly provides the model with target position information without leaking target content, but that operates within a single attention stream. If such a method performs competitively with masked diffusion, it would suggest that position-content separation is the primary bottleneck. If it does not, the failure must stem from something deeper --- namely, the tradeoff we have described.

\subsection{Rotary Position Embeddings}

Rotary Position Embeddings (RoPE) \citep{su2024roformer} encode position information by applying a rotation matrix $R_\theta(m)$ to query and key vectors based on their position index $m$. The standard attention score between positions $i$ and $j$ becomes:
\begin{equation}
    \text{Attn}(i, j) = \frac{(R_\theta(i)\mathbf{q}_i)^\top (R_\theta(j)\mathbf{k}_j)}{\sqrt{d}}
\end{equation}
where $R_\theta(m)$ is a block-diagonal matrix of 2D rotations, rotating each pair of dimensions $(2l, 2l+1)$ by angle $m \cdot \theta_l$ with $\theta_l = 10000^{-2l/d}$. Because the dot product of two rotated vectors depends only on the difference in rotation angles, RoPE naturally encodes relative position:
\begin{equation}
    (R_\theta(i)\mathbf{q})^\top (R_\theta(j)\mathbf{k}) = \mathbf{q}^\top R_\theta(j - i) \mathbf{k}
\end{equation}
This aligns well with the semantic locality bias of language: tokens that are closer together in position receive higher attention scores, all else being equal.

\subsection{Decoupling Keys and Queries}

In any-order autoregression, each structural position $t$ is associated with two semantically meaningful positions: the semantic position of the token currently occupying slot $t$, and the semantic position of the token the model must predict next. We refer to these as the \textbf{lagging position} and the \textbf{leading position}, respectively.

Concretely, let $\pi$ be the permutation defining the generation order. At structural position $t$:
\begin{itemize}
    \item The \textbf{lagging position} $p^K_t = \pi(t)$ is the semantic position of the token at structural slot $t$. This corresponds to \emph{where the token at this position is} in the original sequence.
    \item The \textbf{leading position} $p^Q_t = \pi(t+1)$ is the semantic position of the token the model will predict next. This corresponds to \emph{where the model needs to predict} in the original sequence.
\end{itemize}

The key insight is that these two positions serve different roles in attention. Key vectors represent the information contained at a position, so it is natural to rotate them by the lagging position. Query vectors represent what information is being sought, so it is natural to rotate them by the leading position. This yields the decoupled RoPE attention score:
\begin{equation}
    \text{Attn}(s, t) = \frac{(R_\theta(p^Q_s) \mathbf{q}_s)^\top (R_\theta(p^K_t) \mathbf{k}_t)}{\sqrt{d}}
\end{equation}
Expanding via the relative position property of RoPE:
\begin{equation}
    (R_\theta(p^Q_s)\mathbf{q}_s)^\top (R_\theta(p^K_t)\mathbf{k}_t) = \mathbf{q}_s^\top R_\theta(\pi(t) - \pi(s+1)) \mathbf{k}_t
\end{equation}

The resulting attention score is modulated by the semantic distance between the token at position $t$ and the target position that $s$ is trying to predict. Positions holding tokens that are semantically close to the prediction target naturally receive higher attention.

\subsection{What Decoupled RoPE Solves and What It Cannot}

Decoupled RoPE provides the model with information about both the semantic position of each observed token (through the key rotations) and the semantic position of the prediction target (through the query rotations), without revealing the content at the target position. This directly addresses the position-content separation problem.

However, Decoupled RoPE operates on a single attention computation. The attention distribution at each position must still simultaneously serve both prediction and summarization. 
Moreover, by rotating queries according to the leading position, Decoupled RoPE explicitly biases attention toward positions that are \emph{semantically} close to the prediction target. 
While this is beneficial for prediction, it comes at the cost of attending to \emph{structurally} recent positions that carry the most complete sequence summaries. 
The structural-semantic tradeoff is not resolved --- if anything, it is sharpened, since the RoPE bias actively favors one side of the tradeoff.

This makes Decoupled RoPE a particularly informative probe: it isolates position-content separation as a solved problem, allowing any remaining performance gap relative to masked diffusion to be attributed to the structural-semantic tradeoff.
\section{Experiments}
\label{sec:experiments}

\begin{figure}[t]
    \centering
    \includegraphics[width=\textwidth]{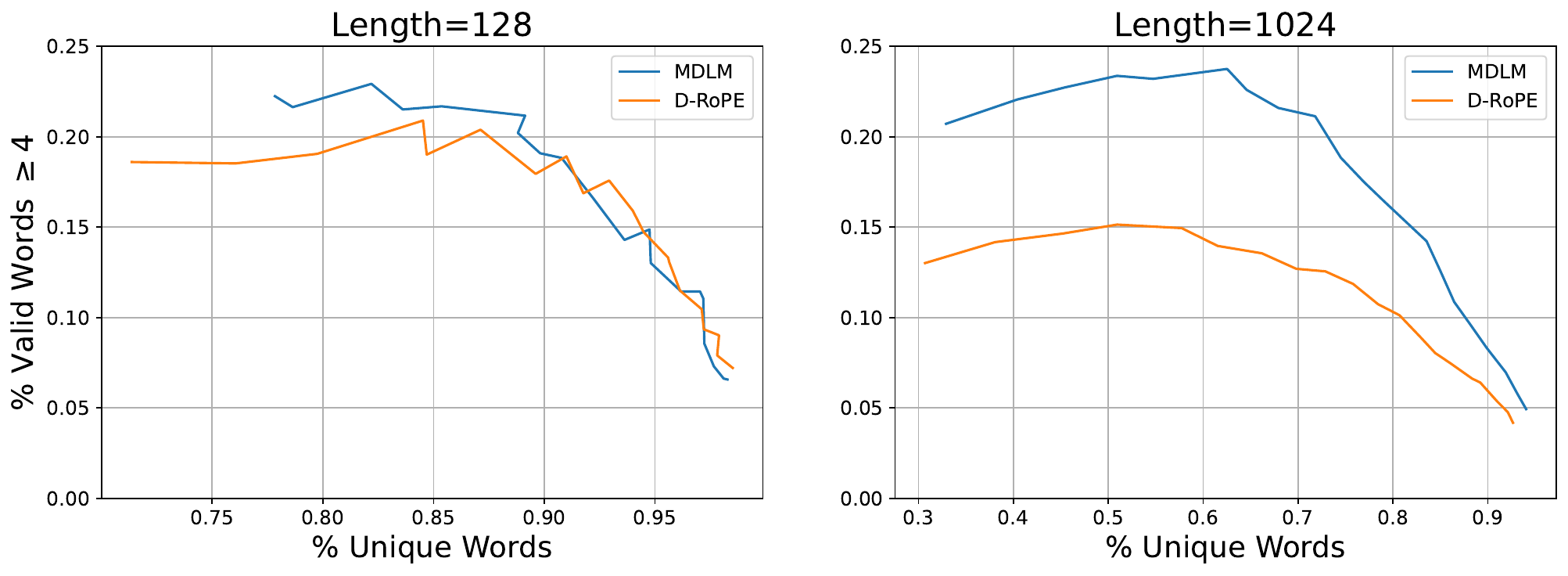}
    \caption{Coherence-diversity frontiers for MDLM and D-RoPE at sequence lengths 128 (left) and 1024 (right). Coherence is measured as the fraction of valid words with four or more characters; diversity is the fraction of unique words. At length 128, the frontiers largely overlap. At length 1024, a clear gap emerges, indicating that D-RoPE's degradation is concentrated in longer words that require global context.}
    \label{fig:frontier}
\end{figure}

\begin{figure}[t]
    \centering
    \includegraphics[width=\textwidth]{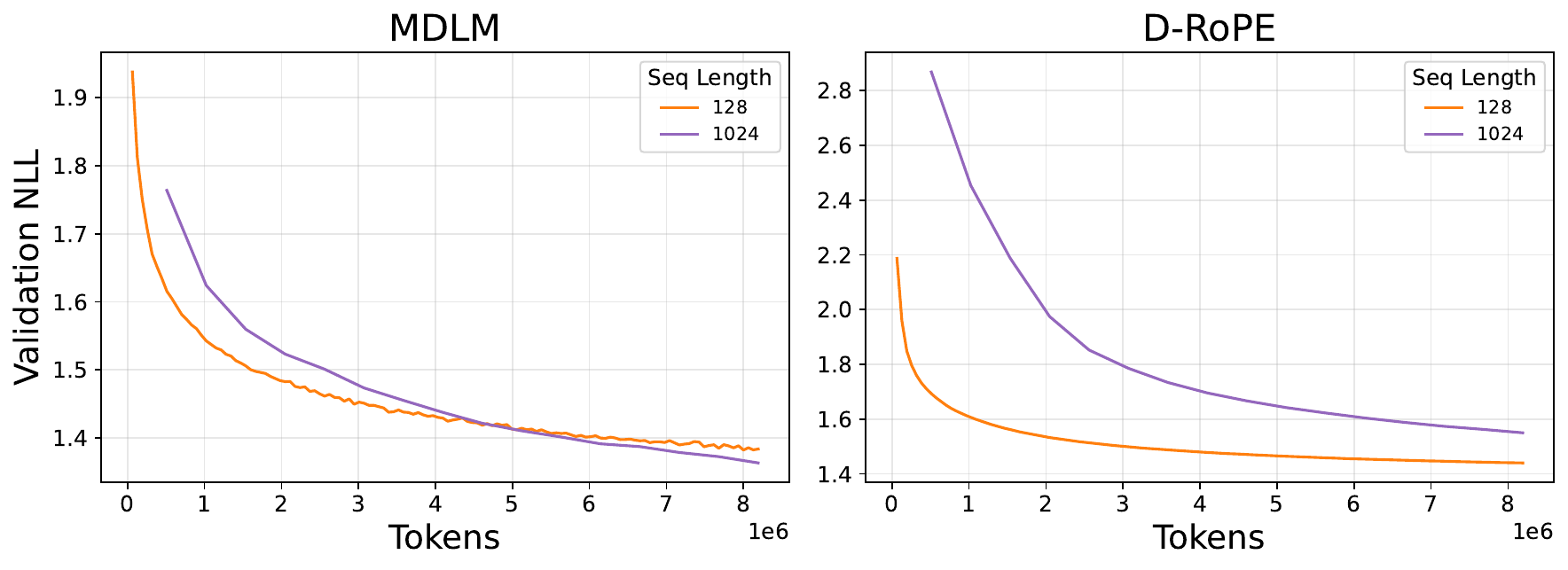}
    \caption{Validation NLL over training for MDLM (left) and D-RoPE (right) at sequence lengths 128 and 1024. For MDLM, the gap between lengths is modest. For D-RoPE, the 1024 curve converges to a substantially worse value, confirming that the length-dependent degradation is specific to the any-order autoregressive parameterization.}
    \label{fig:nll}
\end{figure}

We evaluate Decoupled RoPE on the text8 dataset as a controlled testbed, comparing against masked diffusion. Our experiments are designed to establish two claims: (1) Decoupled RoPE successfully separates position from content and is competitive at short sequence lengths where the structural-semantic tradeoff is mild, and (2) it degrades at longer sequence lengths where the tradeoff becomes severe.

\subsection{Setup}

We train small transformer models (8 million parameters) on text8, a character-level dataset derived from Wikipedia. We compare the following methods:
\begin{itemize}
    \item \textbf{Masked Diffusion (MDLM)}: a bidirectional encoder trained with the standard masked diffusion objective.
    \item \textbf{Decoupled RoPE (D-RoPE)}: a causal decoder trained with the any-order autoregressive objective, using the decoupled RoPE formulation described in the previous section.
\end{itemize}

We evaluate generation quality using \emph{coherence-diversity frontiers}: for each model, we generate sequences at varying temperatures and measure coherence as the fraction of valid English words with four or more characters per generation and diversity as the fraction of unique words per generation. 
We focus on longer words because they are unlikely to appear as valid by chance and typically require coordinating information across a wider context window to produce correctly. 
We report frontiers at sequence lengths 128 and 1024 to contrast behavior when the structural-semantic tradeoff is mild versus severe.

\subsection{Competitive Performance at Short Sequence Lengths}

\Cref{fig:frontier} (left) shows the coherence-diversity frontier at sequence length 128. Decoupled RoPE achieves a frontier that is competitive with MDLM --- the two curves largely overlap across the entire diversity range. 
This indicates that the decoupled formulation successfully provides the model with target position information without leaking content: if the leading and lagging rotations were failing to convey position or were revealing the target token, the model would not achieve a competitive frontier.

\subsection{Degradation at Longer Sequence Lengths}

\Cref{fig:frontier} (right) shows the same comparison at sequence length 1024. Here, a substantial gap emerges: MDLM consistently dominates D-RoPE across the entire frontier. This is consistent with the structural-semantic tradeoff: in longer sequences, the expected structural distance between semantically adjacent tokens grows, and the two orderings become increasingly misaligned. A single attention computation biased toward semantically nearby positions increasingly sacrifices access to structurally recent positions that carry the most complete sequence summaries. 
Because longer words require more global coordination to produce correctly, this tradeoff manifests as a sharp decline in D-RoPE's ability to generate them.

The validation NLL curves in \Cref{fig:nll} corroborate this finding. For MDLM, the the curve for sequence length 1024 converges fairly quickly to the 128 curve. For D-RoPE, the 1024 curve starts substantially higher and converges much slower than the 128 curve. This confirms that the length-dependent degradation is specific to D-RoPE and is not simply a general property of modeling longer sequences.
\section{Conclusion}

We have identified a structural-semantic tradeoff inherent to single-stream causal attention in any-order autoregression. In left-to-right generation, attending to semantically informative tokens and attending to structurally recent tokens are naturally aligned; in any-order generation, these objectives conflict, forcing a single attention mechanism to compromise between local prediction accuracy and global sequence coherence. 

To isolate this tradeoff from the simpler problem of position-content separation, we proposed Decoupled RoPE, which provides target position information without leaking target content. Decoupled RoPE matches masked diffusion at short sequence lengths but degrades as sequences grow longer and the two orderings diverge—precisely the regime where the structural-semantic tension is most acute. This suggests that the success of two-stream attention in prior work stems not merely from separating position from content, but from maintaining separate attention computations that circumvent this deeper tradeoff. We hope this perspective guides future work on efficient parameterizations for any-order autoregressive models.

\paragraph{Limitations.} Our analysis is based on small-scale experiments on a character-level dataset, and the structural-semantic tradeoff is presented as a hypothesis supported by consistent empirical evidence rather than a formal proof. We do not directly evaluate two-stream attention, and it remains possible that other architectural solutions could address the tradeoff without requiring two attention streams. Scaling these experiments to larger models and token-level language modeling would strengthen the conclusions.

\appendix

\bibliographystyle{plainnat}
\bibliography{refs}

\end{document}